\documentclass{article}

\usepackage[utf8]{inputenc} 
\usepackage[T1]{fontenc}    
\usepackage{hyperref}       
\usepackage{url}            
\usepackage{booktabs}       
\usepackage{amsfonts}       
\usepackage{nicefrac}       
\usepackage{microtype}      
\usepackage{xcolor}         
\usepackage{graphicx}       
\usepackage{amsmath}        
\usepackage{array}          
\usepackage{multicol}       
\usepackage{floatrow}       
\usepackage{natbib}         
\title{A Novel Approach for Predicting the Air Quality Index of Megacities through Attention-Enhanced Deep Multitask Spatiotemporal Learning}

\author{%
  \textbf{Harun Khan}\thanks{All authors contributed equally to this research}\\
  W.T. Woodson High School \\
  \texttt{0009-0007-5834-7751} \\
  \and
  \textbf{Joseph Tso}\footnotemark[1]\\
  George Mason University \\
  \texttt{0000-0001-5279-3557} \\
  \and
  \textbf{Nathan Nguyen}\footnotemark[1]\\
  W.T. Woodson High School \\
  \texttt{0009-0003-2048-927X} \\
  \and
  \and
  \textbf{Nivaan Kaushal}\footnotemark[1]\\
  Thomas Jefferson HSST \\
  \texttt{0009-0000-3708-1834} \\
  \and
  \textbf{Ansh Malhotra}\footnotemark[1]\\
  Thomas Jefferson HSST \\
  \texttt{0009-0002-5667-3725} \\
  \and
  \textbf{Nayel Rehman}\footnotemark[1]\\
  Thomas Jefferson HSST \\
  \texttt{0009-0008-5104-064X} \\
}

\begin{document}

\maketitle

\begin{abstract}
Air pollution remains one of the most formidable environmental threats to human health globally, particularly in urban areas, contributing to nearly 7 million premature deaths annually. Megacities, defined as cities with populations exceeding 10 million, are frequent hotspots of severe pollution, experiencing numerous weeks of dangerously poor air quality due to the concentration of harmful pollutants. In addition, the complex interplay of factors makes accurate air quality predictions incredibly challenging, and prediction models often struggle to capture these intricate dynamics. To address these challenges, this paper proposes an attention-enhanced deep multitask spatiotemporal machine learning model based on long-short-term memory networks for long-term air quality monitoring and prediction. The model demonstrates robust performance in predicting the levels of major pollutants such as sulfur dioxide and carbon monoxide, effectively capturing complex trends and fluctuations. The proposed model provides actionable information for policymakers, enabling informed decision making to improve urban air quality.
\end{abstract}

\section{Introduction}
During the past century, exponential population growth and industrialization have significantly threatened urban air quality, affecting the health of billions worldwide. As we face a future of increased urbanization and population growth, exposure to harmful airborne chemicals and particles is expected to worsen \citep{bikis2023urban}. Predicting the Air Quality Index (AQI) and the concentration of airborne pollutants, both in the short-term and long-term, will become crucial. Emerging solutions rely on advanced computational techniques, including machine learning (ML) and deep neural networks (DNN), to analyze patterns in large data sets, allowing cities to better understand and predict the complex interplay of factors influencing their air quality. However, a review of existing studies reveals that many researchers focus on short-term prediction (1 to 3 weeks), with very few examining how long-term (>1 year) urban and climatic factors can impact the average air quality of a city.

In recent studies, significant advances have been made in air quality prediction models. Sun et al. (2020) \citep{sun2021deep} employed a spatial-temporal deep multitask learning approach to forecast short-term AQI in Chinese cities, achieving an accuracy of 87.23\%. Similarly, Hossain et al. (2020) \citep{hossain2020novel} introduced a hybrid deep learning model that combined Gated Recurrent Unit (GRU) and Long Short-Term Memory (LSTM) networks for AQI prediction in Bengali cities, achieving a RMSE of 0.07172. Emeç et al. (2023) \citep{emecc2024novel} developed a stacking ensemble model that achieved 91\% precision in predicting short-term levels of PM2.5. Furthermore, Zhang et al. (2023) \citep{zhang2023prediction} proposed a hybrid model integrating the decomposition of Singular Spectrum Analysis (SSA), Bidirectional Long Short-Term Memory (BiLSTM) prediction and Light Gradient Boosting Machine (LightGBM), achieving an accuracy rate of 99\% in short-term prediction. Although all of these studies demonstrate high accuracy for hourly or daily predictions, none of them discuss the possibility of long-term predictions research or the interpretability of the models.

Long-term prediction and interpretation of urban air quality is essential for informed and impactful decision making by scientists and policy makers \citep{zhang2012real, craig2008air, kirchhoff2013actionable}. Using advanced DNN algorithms for predictive intelligence, decision makers can gain valuable insight into future scenarios. This study aims to achieve three primary objectives. First, it proposes the development of a multitask spatiotemporal learning model as a fundamental tool to predict trends in the Air Quality Index (AQI) in urban areas. Second, it focuses on forecasting urban air quality over extended periods. Third, it aims to help and prepare policymakers to make calculated decisions to support Earth’s air quality.

\section{Materials and Methods}

\subsection{Datasets}

To develop a robust prediction model, it is essential to ensure that the selected data exhibits a meaningful relationship and is captured with high spatial and temporal resolution. In this study, data was sourced from two primary repositories: the Geostationary Operational Environmental Satellites (GOES) operated by the National Oceanic and Atmospheric Administration (NOAA), which collects diverse meteorological measurements, and the Air Quality System (AQS) managed by the Environmental Protection Agency (EPA), which aggregates air quality data from both local and national monitoring stations. The selected data is from the city of Los Angeles, California. The details of the selected variables are provided in Table 1, synthesized from a thorough review of the past literature \citep{sachdeva2024integrated, zhang2018important, ji2020spatiotemporal}. Each variable was fully preprocessed, including resampling to daily averages, to enhance model generalization, convergence, and performance. Key processing steps involved time alignment, interpolation, and imputation to ensure temporal consistency for model training, as detailed below.

\textbf{Interpolation:}
Interpolation is used to approximate missing values within the range of known data points using an estimation function \citep{bergh2012interpolation, davis1975interpolation, gardner1993interpolation}. Linear interpolation was used in this study, described in Equation 1:
\begin{equation}
x(t) = x_i + \frac{(t - t_i)}{(t_{i+1} - t_i)}(x_{i+1} - x_i)
\end{equation}
where \( t_i \) and \( t_{i+1} \) are the known data points before and after \( t \), the desired data point.

\textbf{Imputation:}
Imputation is another commonly used process used to minimize missing data within large datasets \citep{donders2006gentle, rubin2018multiple, schafer1999multiple}. This study used mean imputation which replaces missing values with the mean of the available data to minimize deviation, described in Equation 2:
\begin{equation}
x_{\text{missing}} = \frac{1}{N} \sum_{i=1}^{N} x_i
\end{equation}
where \( N \) is the number of available data points. 

\begin{table}[h]
  \caption{Summary of meteorological and air quality variables}
  \label{summary-of-meteorological-and-air-quality-variables}
  \centering
  \begin{tabular}{lll}
    \\ \toprule
    \cmidrule(r){1-3}
    Variable & Data Source & Units \\
    \midrule
    Temperature & NOAA & C \\
    Wind Speed & NOAA & m/s \\
    Wind Direction & NOAA & degrees \\
    Relative Humidity & NOAA & \% \\
    Precipitable Water & NOAA & cm \\
    Pressure & NOAA & mbar \\
    O$_3$ & EPA & ppm \\
    CO & EPA & ppm \\
    SO$_2$ & EPA & ppb \\
    NO$_2$ & EPA & ppb \\
    \bottomrule
  \end{tabular}
\end{table}

\subsection{Model Architecture and Training}

This study proposes a sophisticated neural network architecture that combines attention mechanisms with long-short-term memory (LSTM) networks. The model is designed to process a multivariate time series input consisting of the 6 input variables in Table 1. The input layer feeds into an attention layer, which allows the model to focus on crucial aspects of the input sequence. Following the attention layer, the processed data is forwarded to an LSTM layer configured with 512 hidden units and 1 layer. To further enhance the performance of the model during training, Batch Normalization (BatchNorm) and Dropout layers are integrated after the first LSTM layer. BatchNorm ensures that the inputs to subsequent layers are consistently normalized, which accelerates training convergence and reduces the likelihood of overfitting. Regularization of dropouts helps to prevent the network from relying too heavily on specific nodes during training. The output from the BatchNorm layer is then processed by a second LSTM layer with 512 hidden units and 1 layer. After the second LSTM layer, another BatchNorm layer is applied to normalize the outputs before they are passed to a linear layer. The linear layer serves as the final stage of the model, with the predicted AQI time series as the output. 

\begin{figure}[!htpb]
    \centering
    \caption{Hyperparameter Settings}
    \begin{tabular}{>{\raggedright\arraybackslash}m{3.5cm} >{\raggedright\arraybackslash}m{2.5cm}}
        \toprule
        Hyperparameter & Specific Settings \\
        \midrule
        Number of iterations & 200 \\
        Learning rate & 0.001 \\
        Loss function & MAE \\
        Batch size & 8 \\
        Input dimension & $6 \times 730$ \\
        Output dimension & $2 \times 730$ \\
        Encoder layers & 2 \\
        \bottomrule
    \end{tabular}
\end{figure}

\section{Results and Analysis}

\subsection{Statistical Analysis}
The learning model was evaluated using various performance metrics, including mean absolute error (MAE), root mean squared error (RMSE), mean squared error (MSE) and mean absolute percentage error (MAPE). These metrics were calculated for four major pollutants: sulfur dioxide (SO$_{2}$), nitrogen dioxide (NO$_{2}$), tropospheric ozone (O$_{3}$) and carbon monoxide (CO). The results are presented in Table 2.

The model demonstrates robust performance in predicting CO and SO$_{2}$ levels, indicated by low MAE and RMSE values. This suggests high precision and accuracy for these pollutants. The higher MAE and RMSE values for NO$_{2}$ and tropospheric O$_{3}$ indicate higher prediction errors, likely due to the increased variability and complexity in the behavior of these pollutants. Despite this, the low MAPE values (NO$_{2}$: 2.66\%, O$_{3}$: 1.55\%) suggest that the model performs well in capturing the overall trend and fluctuations of these pollutants over time.
\begin{table}
\caption{Model performance metrics}
  \label{model-performance-metrics}
  \centering
  \begin{tabular}{lllll}
    \\ \toprule
    \cmidrule(r){1-2}
    Pollutant & MAE & RMSE & MSE & MAPE \\
    \midrule
    SO$_2$ & 0.363 & 0.464 & 0.216 & 7.82\%     \\
    NO$_2$ & 8.025 & 10.331 & 106.739 & 2.66\%      \\
    O$_3$ & 15.350 & 20.513 & 420.781 & 1.55\%  \\
    CO & 1.933 & 2.413 & 5.824 & 2.14\% \\
    \bottomrule
  \end{tabular}
\end{table}

\begin{figure}[h]
    \centering
    \begin{minipage}[b]{0.24\textwidth}
        \centering
        \includegraphics[width=\textwidth]{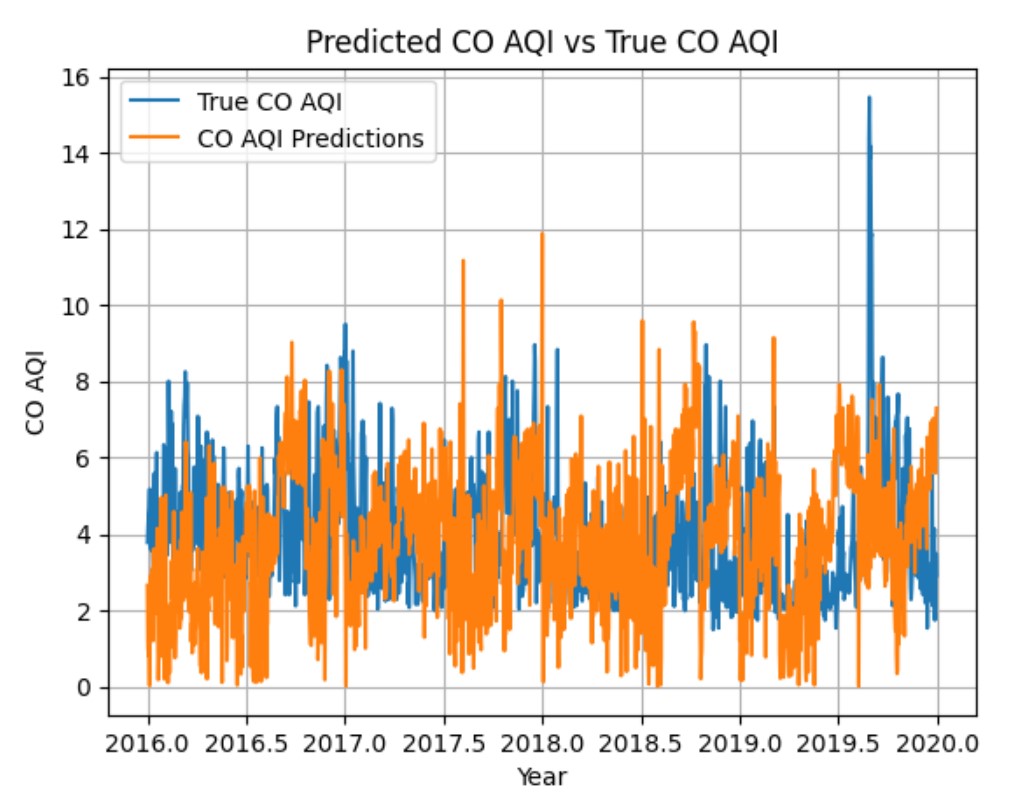}
    \end{minipage}
    \hfill
    \begin{minipage}[b]{0.24\textwidth}
        \centering
        \includegraphics[width=\textwidth]{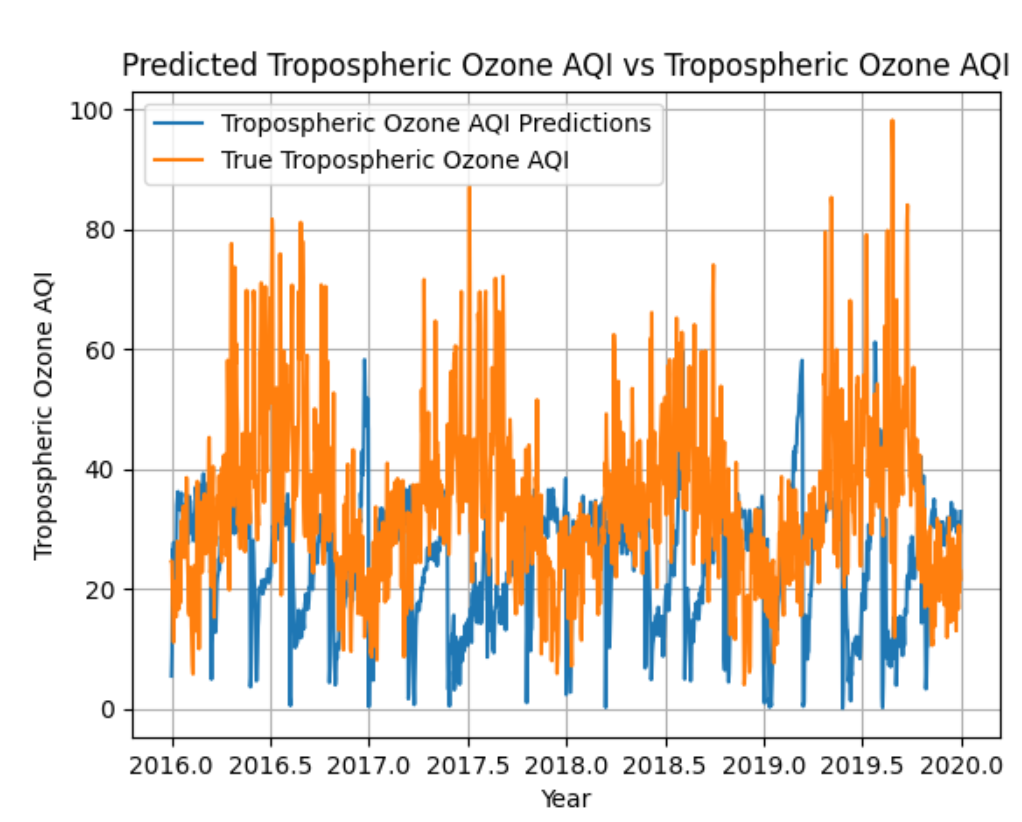}
    \end{minipage}
    \hfill
    \begin{minipage}[b]{0.24\textwidth}
        \centering
        \includegraphics[width=\textwidth]{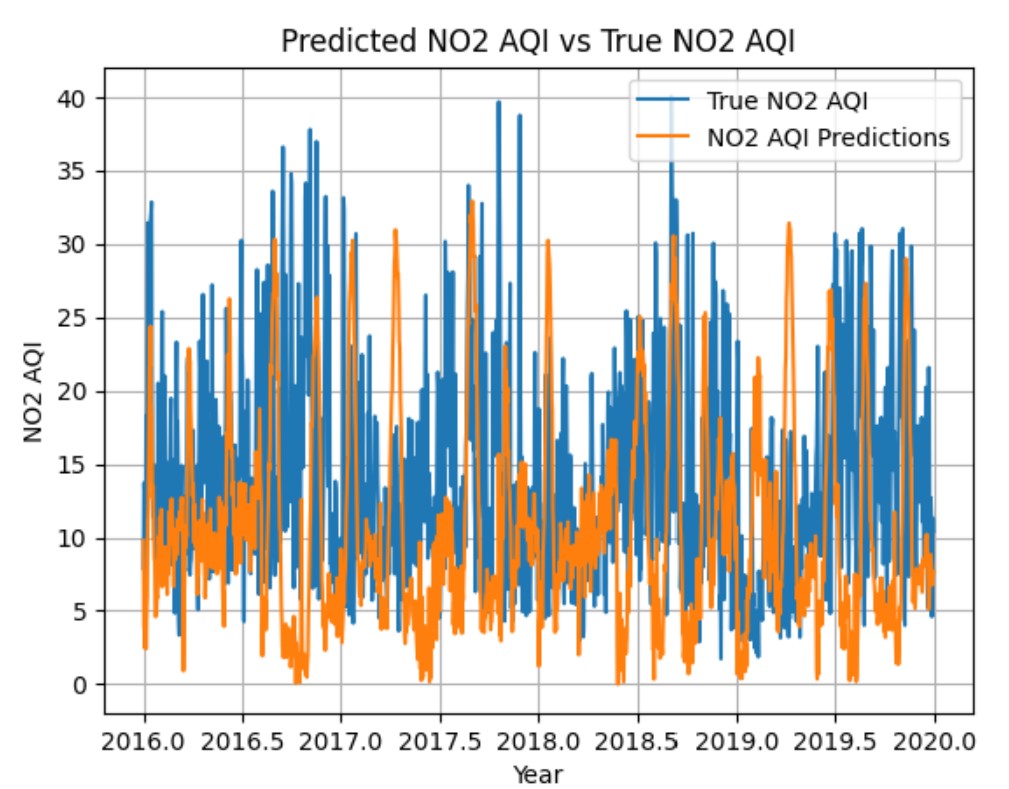}
    \end{minipage}
    \hfill
    \begin{minipage}[b]{0.24\textwidth}
        \centering
        \includegraphics[width=\textwidth]{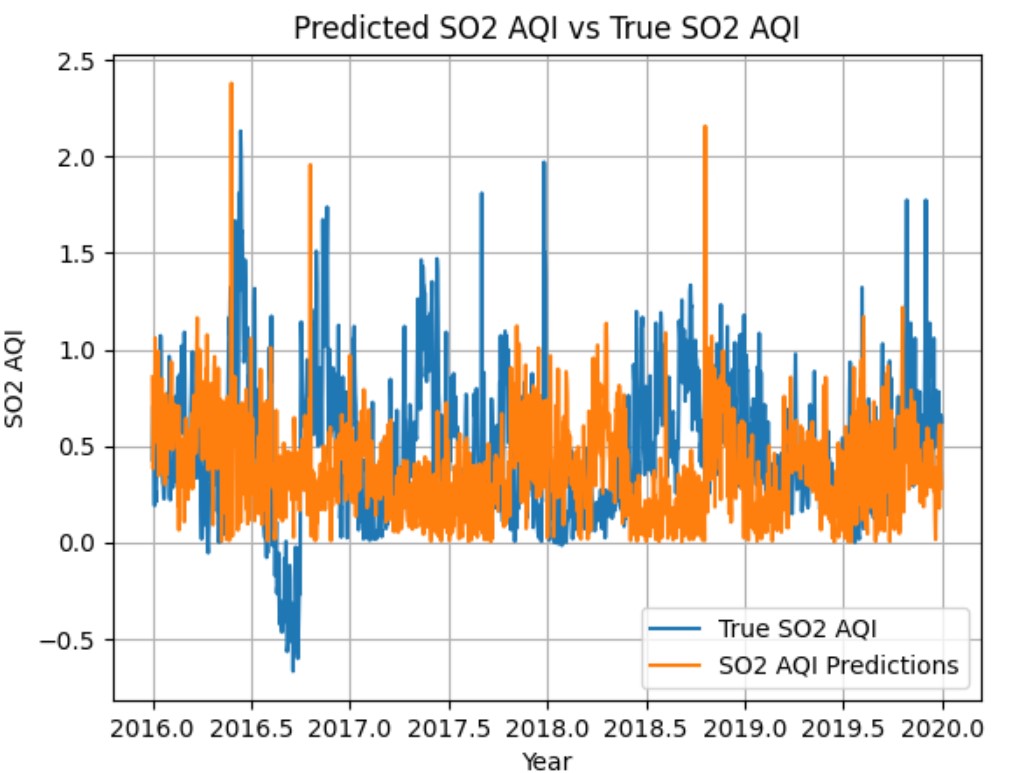}
    \end{minipage}
    \caption{Prediction results}
    \label{fig:single_row}
\end{figure}

\subsection{Feature Importance and Interpretation}
Feature importance for each variable and model was calculated by randomly shuffling the variable column and considering the MAE. If the MAE varied negatively from the base MAE, then the feature is considered important. Based on Figure 3, we see that pressure and relative humidity are consistently important in predicting pollution values. Furthermore, we see that CO is strongly influenced by relative humidity. This type of information can be invaluable to researchers or governments looking to understand how to combat or influence air pollution through the characteristics of the weather.
\begin{figure}[h]   
    \centering
    \begin{minipage}[b]{0.22\textwidth}
        \centering
        \includegraphics[width=\textwidth]{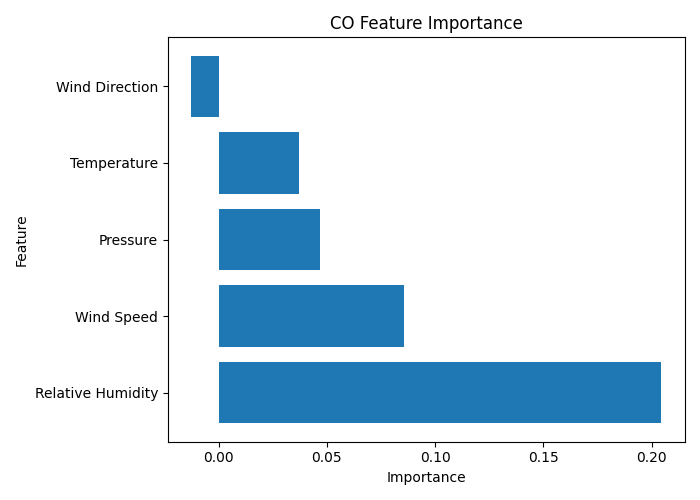}
    \end{minipage}
    \hfill
    \begin{minipage}[b]{0.22\textwidth}
        \centering
        \includegraphics[width=\textwidth]{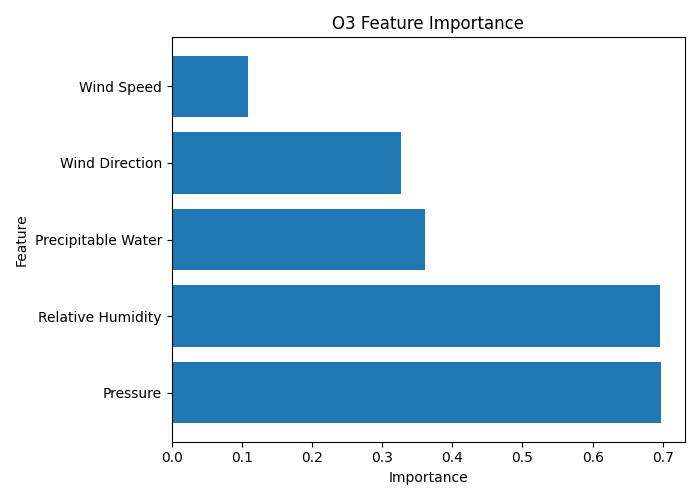}
    \end{minipage}
    \hfill
    \begin{minipage}[b]{0.22\textwidth}
        \centering
        \includegraphics[width=\textwidth]{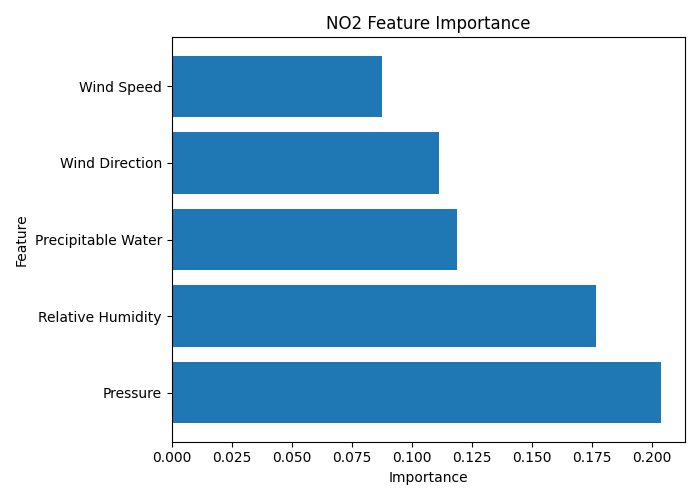}
    \end{minipage}
    \hfill
    \begin{minipage}[b]{0.22\textwidth}
        \centering
        \includegraphics[width=\textwidth]{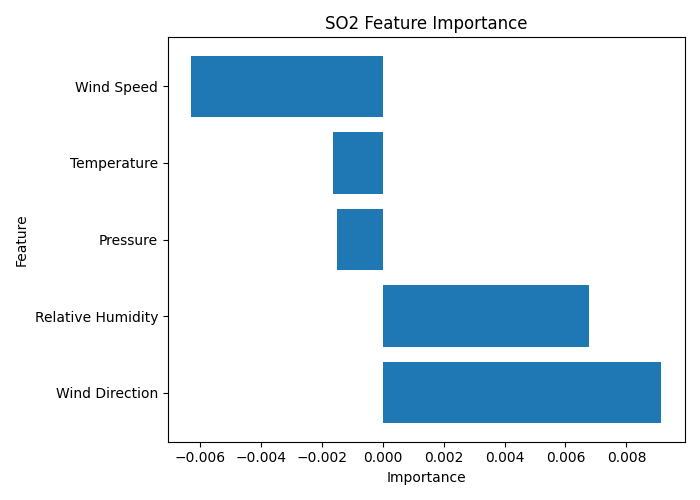}
    \end{minipage}\\
    \caption{Variable importance results}
    \label{fig:single_row}
\end{figure}

\section{Conclusion}
This paper proposed an attention-enhanced multitask spatiotemporal machine learning approach to air quality prediction in megacities. This model is able to achieve high accuracy in forecasting pollutants like nitrogen dioxide and carbon monoxide, but also capture long-term trends in urban air pollution within megacities. Furthermore, through the use of deep learning interpretation techniques, the framework allows for extensive insight into urban air quality prediction, aiding policymakers to mitigate air pollution in the aforementioned megacities. Furthermore, the integration of spatiotemporal data enhances the model's environment, making it adaptable in a world of growing urban development. The combination of interpretability and machine learning methods in this model marks a notable advance in air quality prediction.

Future implications of this project may include additional environmental factors, as well as population density, as factors that influence the prediction of the air quality index in large megacities for more nuanced predictions. This research represents a significant advancement in air quality index prediction, demonstrating the potential of machine learning in long-term prediction models for urban air quality.

\bibliography{citations}

\end{document}